\documentclass{article}

\usepackage{arxiv}

\usepackage[utf8]{inputenc} % allow utf-8 input
\usepackage[T1]{fontenc}    % use 8-bit T1 fonts
\usepackage{hyperref}       % hyperlinks
\usepackage{url}            % simple URL typesetting
\usepackage{booktabs}       % professional-quality tables
\usepackage{amsmath}       % blackboard math symbols
\usepackage{nicefrac}       % compact symbols for 1/2, etc.
\usepackage{microtype}      % microtypography
\usepackage[numbers]{natbib}
\usepackage{graphicx}
\usepackage{soul}           % highlighting
\usepackage{xcolor}
\sethlcolor{yellow}
\definecolor{gpt4}{rgb}{0.79, 0.64, 0.99}
\graphicspath{ {./images/} }
\usepackage{longtable}
\usepackage{pdflscape}
\usepackage{geometry}
\usepackage{booktabs}
\usepackage{array}
\title{Examining the Potential and Pitfalls of ChatGPT in Science and Engineering Problem-Solving}

\author{
 Karen D. Wang \\
  Graduate School of Education\\
  Stanford University\\
  \texttt{kdwang@stanford.edu} \\
   \And
 Eric Burkholder \\
  Department of Physics\\
  Auburn University\\
  \texttt{ewb0026@auburn.edu} \\
  \And
 Carl Wieman \\
  Department of Physics, \\
  Graduate School of Education\\
  Stanford University\\
  \texttt{cwieman@stanford.edu} \\
  \And
  Shima Salehi \\
  Graduate School of Education\\
  Stanford University\\
  \texttt{salehi@stanford.edu} \\
  \And
  Nick Haber \\
  Graduate School of Education\\
  Stanford University\\
  \texttt{nhaber@stanford.edu} \\
}

\begin{document}
\maketitle
\begin{abstract}
The study explores the capabilities of OpenAI's ChatGPT in solving different types of physics problems. ChatGPT (with GPT-4) was queried to solve a total of 40 problems from a college-level engineering physics course. These problems ranged from well-specified problems, where all data required for solving the problem was provided, to under-specified, real-world problems where not all necessary data were given. Our findings show that ChatGPT could successfully solve 62.5\% of the well-specified problems, but its accuracy drops to 8.3\% for under-specified problems. Analysis of the model's incorrect solutions revealed three distinct failure modes: 1) failure to construct accurate models of the physical world, 2) failure to make reasonable assumptions about missing data, and 3) calculation errors. The study offers implications for how to leverage LLM-augmented instructional materials to enhance STEM education. The insights also contribute to the broader discourse on AI's strengths and limitations, serving both educators aiming to leverage the technology and researchers investigating human-AI collaboration frameworks for problem-solving and decision-making. 
\end{abstract}

% keywords can be removed

\section{Introduction}
The rapid advancement of Large Language Models (LLMs) has attracted substantial attention from both the general public and academia. LLMs, such as GPT-4 by OpenAI, can generate human-like textual responses to text-based queries in real-time. Since the public launch of ChatGPT in November 2022, there has been a growing body of research exploring its various capabilities, limitations, and implications across diverse discipline and tasks. One such field is education, where LLMs have far-reaching implications for both instructional practices, \textbf{\textit{how}} we teach and assess; as well as for curriculum content, \textbf{\textit{what}} we teach and assess. 

\par Within the education context, we studied ChatGPT’s capacity for solving problems from a college-level engineering physics course. ChatGPT by OpenAI is one of most accessible and publicly used LLM-based tools, and its most advanced underlying model to date is GPT-4. GPT-4 has outperformed previous models like GPT-3 in an array of standardized exams in disciplines such as law and medicine. Notably, it has achieved scores in the 66th to 84th percentile on the AP Physics 2 Exam \cite{openai2023gpt4}, which features problems that are mostly situated in abstract scenarios and provide all necessary data in the problem statement. However, the literature has so far offered limited insights into the capability of GPT-4 in solving problems that are in real-world contexts and/or do not provide all the data needed for reaching a solution. Consequently, the nuances of GPT-4 problem-solving capability, including the range of problems that it can effectively solve and the quality of the generated solutions, remain largely unknown. 

\par Investigating GPT-4’s problem-solving capabilities has multifaceted implications that extend from enhancing educational practices to fostering human-AI collaboration. First, a more nuanced understanding of how GPT-4 solves different types of problems can offer insights into how to design LLM-augmented instructional materials to support student problem-solving. In this study, we focused our attention on scientific problem-solving with the long-term goal of leveraging LLM-based tools to enhance Science, Technology, Engineering, and Mathematics (STEM) education. This focus stems from the recognition that, despite problem-solving being widely acknowledged as a fundamental learning goal in STEM education \cite{RN21}, effective ways to teach problem-solving remain elusive and understudied. Second, as students start using LLM-based tools such as ChatGPT for their homework problems they need help with \cite{shoufan2023exploring}, they need to be educated about its affordances and limitations to make effective use of such tools for their own learning. Furthermore, beyond its value in educational settings, knowledge of GPT-4's problem-solving capabilities contribute to the broader discourse on human-AI collaboration. Understanding the areas where AI excels and where it currently falls short can inform the development of a human-AI collaborative problem-solving framework.

In this study, we pose the following research questions:
\begin{itemize}
    \item How does ChatGPT’s problem-solving capability vary across different types of physics problems? 
    \item What are ChatGPT's common failure modes for different types of physics problems?
    \item To what extent can standard prompt engineering techniques improve ChatGPT's performance for different types of physics problems?
\end{itemize}

\section{Background}
Human problem-solving has been studied across diverse research traditions and domains, including cognitive psychology, information processing, and discipline-based education research \cite{newell1972human, bransford1986teaching, chi1981expertise, reif1982knowledge}. As different types of problems call for distinct problem-solving strategies and bodies of knowledge, one's problem-solving capability may significantly vary across problem types. Similarly, to illuminate the problem-solving capability of AI models such as GPT-4, we must first explicate the characteristics of the problems given to the models, and study the performance of these models across different problem types. 

\par
\begin{figure}[htp]
  \centering
  \includegraphics[width=0.55\textwidth]{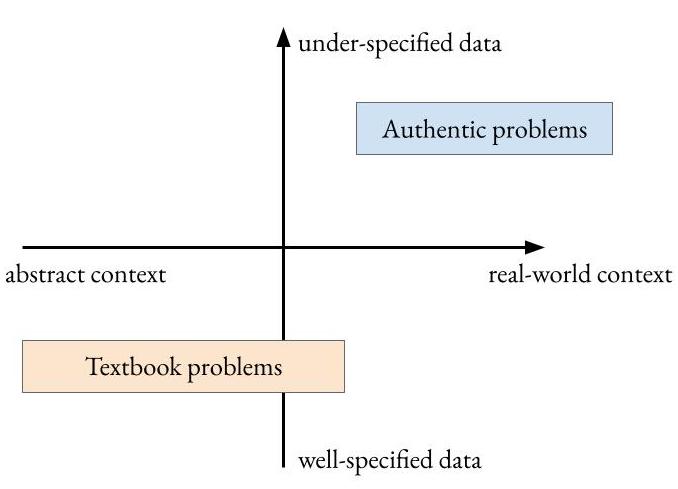}
    \caption{A two-dimensional plane to visualize the problem categorization framework. The x-axis represents the context dimension, ranging from abstract to real-world. The y-axis represents the data specificity dimension and ranges from well-specified to under-specified.}
  \label{fig:framework}
\end{figure}

 
Our research group has done extensive work to characterize and assess authentic problem-solving expertise across science, engineering, and medicine domains \cite{price2021detailed, price2022accurate, salehi2018improving, burkholder2020template}. Drawing on these work, we now characterize problems in science and engineering domains along two dimensions: context and data specificity (Figure~\ref{fig:framework}). The first dimension refers to the context where the problem is situated and spans from abstract to real-world. Abstract problems employ simplified, idealized scenarios that do not exist in the real world, such as frictionless planes and massless pulleys. On the other end of the spectrum are real-world problems that are based on scenarios that individuals may encounter in their daily lives or in professional settings. The second dimension is around the specificity of the data required to solve a problem. Well-specified problems provide all the data required for a solution, while under-specified problems lack some essential data, requiring the problem solver to determine what data is needed and how to obtain it for solving the problem. Textbook problems typically present well-specified data and may have either abstract or real-world context. These problems are designed to make it easier for learners to grasp and practice domain-specific concepts. In contrast, authentic problems bring with them the complexity and ambiguity that comes from real-world challenges and do not specify all the required data. 

\par The above problem categorization framework is intended for analyzing problems that are knowledge-rich, or requiring the application of content knowledge from STEM disciplines. These problems differ from the classic knowledge-lean problems employed to study problem-solving in the information processing paradigm \citep{simon1973structure}. The knowledge-lean problems, such as the Tower of Hanoi, are often termed as "well-defined" to indicate that they have clear initial and goal states and a set of clearly-defined operators for moving from the initial state to the goal state \citep{jonassen1997instructional, simon1978information}. It is important to differentiate "well-defined" and "ill-defined" from the "well-specified" and "under-specified" terminology we used in the problem categorization framework. The former terms capture the clarity of the initial and the goal states of a problem and the constraints on the possible operations to navigate from one to the other, while the latter terms are used for evaluating the quantity and clarity of data given in the problem statement.

\par Most existing research has focused on examining AI’s performance in handling textbook-style problems that are well-specified and mostly abstract. For example, GPT-4 has performed well in standardized tests such as AP Biology, Chemistry, Environmental Science and Physics Exams \cite{openai2023gpt4, nori2023capabilities}. The model also demonstrated proficiency surpassing average human performance in writing program functions that solely depend on existing public libraries \cite{bubeck2023sparks}. In contrast, there is a scarcity of research on how AI approaches authentic problems that are under-specified and situated in real-world contexts, even though such authentic problems are likely to constitute a significant share of the tasks that AI will encounter when deployed in the real world. Emerging research that ventures into the related domain has investigated AI's capacity for inductive reasoning, which involves identifying general principles from a small set of examples and applying these principles to novel situations \citep{wang2023hypothesis, xu2023llms, gendron2023large}. Results of these investigations suggest significant room for improvement in AI's capability to make generalizations from specific instances. 

 % The US Next Generation Science Standards (NGSS) placed an explicit focus for students to learn solving such authentic problems as part of science and engineering practices to prepare for future challenges in a rapidly changing world \cite{national2013next}. Consequently, it is important to examine how AI approaches authentic, under-specified problems to understand its educational implications for supporting students’ development of problem-solving competencies. Moreover, such under-specified problems are likely to constitute a significant share of the tasks that AI will encounter when deployed in the real world, making it imperative to examine its current performance on them. 

% \par What makes a problem authentic? Based on our previous research \cite{price2022accurate}, we consider the following to be important characteristics of authentic problems in science and engineering domains: 1) requiring domain knowledge, 2) providing limited data upfront, 3) prescribing no clear solution path or criteria for success. It should be noted that these represent only a subset of the broader array of characteristics of real-world problems. Nonetheless, the highlighted characteristics of authentic problems can serve as a useful guide for designing assessment tasks to more accurately gauge students' as well as AI models’ problem-solving competencies than typical, well-specified textbook and exam questions.  

\par While GPT-4's performance in solving textbook-style problems should not be extrapolated to its performance on authentic problems, a review of previous literature nonetheless provides insights into some of its common failure modes. One common flaw in GPT-4's performance is related to calculation errors. Previous studies have found that while the model can answer difficult high-school level math questions and discuss advanced mathematics concepts, it could also make basic errors in calculation (e.g., arithmetic mistakes) \cite{bubeck2023sparks}. Another limitation is the model's deficiency in critically evaluating its own solutions. This leads to failure in recognizing mistakes in its solution path\cite{bubeck2023sparks, zhang2023language}. A separate study employed GPT-3.5 and GPT-4 to answer open-domain questions, such as whether the New Orleans Outfall Canals are the same length as the Augusta Canal. The researchers summarized the models' failure modes in solving these problems into four categories: comprehension error, factualness error, specificity error, and inference error \cite{zheng2023does}. The study found that nearly half of the failures were due to factualness error, or the model lacking the necessary supporting facts to produce a correct answer, and another 25\% of the failures were due to inference error, or the model failing to reason effectively. 

\par In the context of physics education, a study reported that ChatGPT (based on the GPT-3 model) could narrowly pass a calculus-based college-level introductory physics course \cite{PhysRevPhysEducRes.19.010132}. One test used for evaluation was the Force Concept Inventory (FCI), which comprises well-specified multiple-choice questions. GPT-3 solved 60\% (18 of 30) of the FCI items. Moreover, the researcher found that the model's performance variation was more influenced by the mathematics than the physics concepts involved. Similar to the above mentioned studies, this study found that ChatGPT had persistent problems with calculation, especially in manipulating and calculating formulas involving square roots. 

\section{Methods}
\subsection{Problems Used in the Study}
A total of 40 homework problems from an engineering physics course taught by one of the authors were used in this study (Appendix \ref{app:problem_details}). The course covers an array of topics including force, torque, linear/rotational/harmonic motion, and fluid mechanics, and aims at developing students’ problem-solving competencies. Based on our proposed problem categorization framework (Fig~\ref{fig:framework}), we characterize these problems along the two key dimensions: context and data specificity. Regarding the context dimension, the problems are all situated in real-world contexts. For example, one problem involves calculating the total travel time for an elevator ascending to the top floor of the Salesforce Tower in San Francisco, and another involves selecting fishing lines that are strong enough to hang sculptures from the ceiling of an atrium of a new building.

Regarding the data specificity dimension, the problems used in this study span a spectrum from well-specified to under-specified. On one end of the spectrum are problems that provide all the data needed for solving, including values for key variables and parameters. On the other end are problems with under-specified or incomplete data, requiring the problem solver to determine what data is needed and how to collect the missing data. This variation in data specificity necessitates different levels of decision-making by the problem solver regarding data collection, which is a key practice for solving authentic problems as identified in our previous research \cite{salehi2018improving, price2021detailed}. By incorporating this range of problem types, we are able to conduct a more comprehensive and nuanced evaluation of ChatGPT’s problem-solving capability.

%The real-world feature of the problems differs from the physics problems employing abstract scenarios that are detached from the complexities of real-world challenges. Second, solving these problems require the application of domain-specific knowledge. Specifically, a problem solver must identify and apply the relevant physics principles and formula to model the situation and arrive at a viable solution. While all problems in this study are situated in real-world contexts and require domain knowledge, 

\par Table \ref{tab:sample_problem} presents two sample problems used in the study. Both problems are situated in real-world contexts. The first one is a well-specified problem where all data needed to solve the problem was provided in the problem statement. In contrast, the second one represents an under-specified problem where the problem statement does not provide any data, and necessitates the problem solver to collect all the required data through conducting an online query or making reasonable assumptions in order to solve the problem. 

\begin{table}[htp]
 \caption{Two sample problems used in the study.}
  \centering
  {\renewcommand{\arraystretch}{1.5}
    \begin{tabular}{p{7.5cm}|p{7.5cm}}
    \hline
    \textbf{The Log Cabin Problem (Well-specified)} & \textbf{The Dresser Tip-over Problem (Under-specified)} \\
    \hline
    You are planning to build a log cabin and will need to pull the logs up a hill to the building site by means of a rope attached to a winch. In order to buy the rope, you need to know how strong the rope must be and decide to do a quick calculation for this. The heaviest of the logs weigh 500 lbs. You estimate the coefficient of friction between the log and the hill to be 0.8, and the hill you have to pull them up is at an angle of 30 degrees. How strong must the rope be? & IKEA has had some issues with children climbing their dressers by pulling on the drawers and getting hurt when the dressers tip over. Their solution has been to provide wall mounts that you can use to secure the top of the dresser to the wall. Figure out how strong this wall mount has to be to keep the dresser from tipping over. You should include an equation showing how your answer depends on the weight of the child and the size and weight of the dresser. \\
    \hline
    \end{tabular}}
  \label{tab:sample_problem}
\end{table}

\subsection{Experiments and Analysis}
We used ChatGPT with GPT-4 selected as the underlying model in the present study. The decision to use ChatGPT as opposed to running the model through OpenAI's API was grounded in the interest of ecological validity, as students and instructors of STEM courses are more likely to use ChatGPT than to access the GPT-4 model directly. This methodological choice allows our study's findings to be more directly applicable to the common STEM educational settings where LLM-based tools are used. 

\par Each problem statement of the 40 problems was pasted into the dialogue interface of ChatGPT, accompanied by the prompt of "solve the following physics problem." No additional guidelines or contextual knowledge was provided. If ChatGPT returned with queries or statements indicating that the problem could not be solved without additional information, a second prompt was put into the dialogue box to direct the model to make reasonable assumptions and solve the problem. Once the model reached a final answer, its response was transferred to a centralized document for record and analysis. This approach was implemented to minimally influence ChatGPT's problem-solving approach and establish a baseline for its problem-solving capability. 

\par To evaluate ChatGPT’s problem-solving capability, the first step was to determine the accuracy of its solutions by examining if it arrived at the correct final answer. In instances where the model failed to come up with the correct solution, we conducted a step-by-step analysis of the solutions it generated against the correct solutions produced by the course’s lead instructor to identify where in the solution process ChatGPT failed. The results of this error analysis were recorded for each individual problem and subsequently categorized into distinct failure modes based on the nature of the error. The goal of this categorization was to identify patterns and recurrent themes in ChatGPT’s errors and to generate insights into its capabilities and limitations. 

\par Next, we examined whether simple prompt engineering could improve ChatGPT's problem-solving performance. In the context of AI research, prompt engineering refers to the process of designing, testing, and refining inputs given to AI models to enhance their performance \cite{liu2021pretrain}. Prompting strategies such as zero-shot chain-of-thought have demonstrated success in improving LLMs' performance in solving multi-step arithmetic word problems by instructing the model to "think step-by-step" \cite{kojima2023large}. In the second phase of the study, we adopted a similar prompting strategy for the problems in our dataset. Specifically, the prompt was updated to "solve the following physics problem step-by-step" just before presenting the problem statement to ChatGPT. The intention was to explore whether ChatGPT could decompose the problem into more manageable sub-problems and circumvent the errors it made during its initial problem-solving attempt.  

\section{Results}
In this section, we first present ChatGPT's problem-solving success rate without the use of prompt engineering. Next, we discuss three distinct failure modes based on a comparative analysis between ChatGPT's incorrect solutions and the instructor's correct solutions. Finally, we explore how prompt engineering impacted ChatGPT's problem-solving performance. 

\subsection{Problem-solving Performance}
Our analysis revealed a substantial difference in the ChatGPT's ability to solve well-specified vs. under-specified types of problems (Table~\ref{tab:performance}). ChatGPT successfully solved 62.5\% of the well-specified problems and only 8.3\% of the under-specified problems. This discrepancy in accuracy rate was statistically significant (Fisher's exact test, \textit{p} \textless{} 0.001). 

\begin{table}[h]
    \caption{ChatGPT's problem-solving performance grouped by the problems' data specificity}
    \label{tab:performance}
    \centering
{\renewcommand{\arraystretch}{1.5}
    \begin{tabular}{lccc}
        \toprule
        Problem Type & Number of Problems & Number Correctly Solved by ChatGPT & Accuracy Rate \\
        \toprule
        Well-specified & 16 & 10 & 62.50\% \\
        Under-specified & 24 & 2 & 8.33\% \\
        \bottomrule
    \end{tabular}}
\end{table}

At the same time, ChatGPT demonstrated a high level of proficiency in identifying the relevant physics concepts to apply based on the given problem statement. This capacity was evidenced by the model's consistent performance of outlining the relevant physics concepts at the beginning of the solutions it generated. ChatGPT's strength in this facet of problem-solving differs from typical human performance, as students often struggle to identify what concepts to apply as the starting point in solving unfamiliar problems. Additionally, students may struggle with complexities that arise in correctly identifying, applying and integrating domain knowledge learned at different times. In contrast, LLMs like GPT-4 are not constrained by such linear learning pathway, and their training data is likely to incorporate a more comprehensive range of domain-specific knowledge compared to what students learn in a typical college-level curriculum. This expansive knowledge base is one of AI's strengths in addressing real-world challenges.   

\subsection{Types of Failure Modes}
A detailed table of all 40 problems used in the study, along with their underlying physics concepts and ChatGPT's performance, can be found in Appendix~\ref{app:problem_details}. In examining where all of ChatGPT's incorrect solutions diverged from the instructor's solutions, we identified three distinct types of failure modes. These failure modes can help us understand the underlying causes of the AI model's difficulties in problem-solving. 

\subsubsection{Failure to construct accurate models of the physical world}
\begin{figure}[htp]
  \centering
  \includegraphics[width=0.5\textwidth]{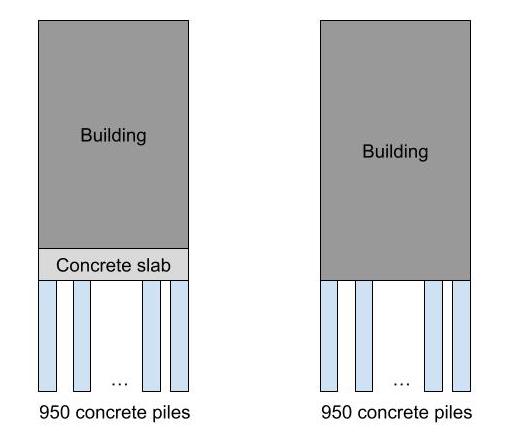}
  \caption{Two contrasting models for the Millennium Tower problem. The first one (left) is the accurate model and the second one (right) was created based on ChatGPT's incorrect solution}
  \label{fig:millenium_tower}
\end{figure}

One of the failure modes is related to ChatGPT's failure to construct accurate models of the physical world based on the problem statement. Table~\ref{tab:millenium-tower} presents a problem from our dataset and ChatGPT's incorrect solution that falls into this type. The problem asks for an estimation of the force required from each of the 950 friction piles to prevent the Millennium Tower in San Francisco from sinking. This is an under-specified problem, as the problem statement does not specify the density of concrete. The highlighted texts in the problem statement are information that ChatGPT did not account for in its modeling of the problem. A human problem solver could intuitively construct a model of the problem, that both the building and the concrete slab are on top of the 950 concrete piles, as depicted by the first model in Figure~\ref{fig:millenium_tower}. This model helps the problem solver focus on the key objects and forces involved. The force required from each friction pile to support the building and concrete slab in addition to its own weight would therefore be: $F_{\text{friction}} = (W_{\text{building}} + W_{\text{slab}}) / 950 + W_{pile}.$
In contrast, the second model in Figure~\ref{fig:millenium_tower} was created by the researcher based on ChatGPT's incorrect solution. ChatGPT failed to consider the weight of the concrete slab and piles in its calculations, indicating that it did not correctly conceptualize the spatial relationship between the building, the concrete slab, and the 950 piles. This led to an oversimplified and erroneous calculation of the force required from each pile.

\begin{table}[htp]
\centering
\caption{The Millennium Tower problem and ChatGPT's solution }
\label{tab:millenium-tower}
{\renewcommand{\arraystretch}{1.5}
\begin{tabular}{p{7cm}|p{8cm}}
\hline
\textbf{The Millennium Tower Problem} & \textbf{A Summary of ChatGPT's Solution} \\
\hline
\sethlcolor{orange}
\hl{Prompt:}
\sethlcolor{yellow}
Solve the following physics problem: The Millennium Tower in San Francisco was completed in 2009 and has received a lot of attention recently due to problems with the building settling and leaning. The building is 605 feet tall, has a base of about 20,000 square feet, and weighs about 7 tons per square foot of its foundation. The tower sits on \hl{a 10-ft thick concrete slab}, which is in-turn supported by 950 friction piles, which are \hl{14-inch square concrete pillars} embedded in the bay sand. \hl{The piles are 80 feet long.} Estimate the approximate force required from each friction pile to keep the building from sinking. & 

\sethlcolor{gpt4} \hl{ChatGPT:} To estimate the force each pile needs to exert, we need to determine the weight of the building and distribute that weight across the number of friction piles.

The total weight of the building is given as 7 tons per square foot of its foundation. The foundation is approximately 20,000 square feet. Therefore, the total weight is 7 tons/ft² * 20,000 ft² = 140,000 tons.

Since 1 ton is equivalent to 2,000 pounds, the total weight of the building is 140,000 tons * 2,000 pounds/ton = 280,000,000 pounds. 

The force each pile needs to exert is 280,000,000 pounds / 950 = 294,737 pounds (approximately), assuming an even distribution of the weight of the building. \\
\hline
\end{tabular}}
\end{table}

\par Out of the 28 problems that ChatGPT did not solve, 14 can be attributed to this failure mode of it not being able to accurately model the physical world, especially regarding spatial relationships. Another example of this failure mode is the Dresser Tip-over problem that asks students to determine the force exerted by a wall mount to prevent a dresser from tipping over when children pull down on the drawers (see the problem in Table~\ref{tab:sample_problem}). Solving the problem requires an accurate model representing the center of mass of the dresser, the pulling by a child, the location of the wall mount, and most importantly, the pivot point and the distances to the pivot point for each of the forces. One of its persistent challenges observed in our analysis is that ChatGPT cannot correctly identify the pivot point. In other words, it cannot correctly conceptualize how the dresser would fall if pulled by a child. 

\subsubsection{Failure to make reasonable assumptions about missing data}
A second failure mode of ChatGPT in problem-solving was its limitations in making reasonable assumptions about missing data in under-specified problems. When presented with problems that lack complete data, ChatGPT usually attempted to fill in the missing data and generate a solution on it own. Only for two under-specified problems did it respond to the initial problem statement by stating that the problem doesn't provide all the necessary information to calculate the exact value for a solution. In such cases, a follow-up instruction was entered into ChatGPT's dialogue box, directing the model to make reasonable assumptions about the specific value it noted and solve the problem. Table~\ref{tab:follow-up} presents such an instance where ChatGPT explicitly listed the missing information. 

\begin{table}[htp]
    \centering
    \caption{An Example of ChatGPT requesting more information for the Floating Duck problem}
    \label{tab:follow-up}
    {\renewcommand{\arraystretch}{1.5}
    \begin{tabular}{p{0.95\linewidth}}
        \hline
        \sethlcolor{orange}
        \hl{Prompt:} Solve the following physics problem: Off-shore oil-drilling can be controversial due to the possibility of oil spills. A sea duck can float on salt water with only about 20\% of its body submerged below the surface. How thick would a layer of crude oil have to be for the duck to drown? For reference, the BP oil spill in 2010 reached a depth of 41.5 meters.\\

        \hline
        \sethlcolor{gpt4}
        \hl{ChatGPT:}
        To solve this problem, we first need to understand the concept of buoyancy. Buoyancy is the force exerted on an object that is wholly or partly immersed in a fluid. In our case, the fluid is the combination of salt water and crude oil. The sea duck floats with only 20\% of its body submerged in the water because the buoyancy force from the submerged part of the duck equals the weight of the duck. This means that the duck is in equilibrium.

    If a layer of oil were to cover the surface of the water, the duck would start to submerge into the oil, since oil is less dense than water. If the oil layer is thick enough, the duck could potentially sink below the surface and drown. To find the thickness of the oil layer that would cause the duck to drown, we need to find the point where the duck's buoyancy in the oil equals its weight. 

\sethlcolor{yellow}
\hl{
Here's where we run into a problem, though. To find this value, we need additional information that isn't provided in the problem: }

1. The volume of the duck. This would give us the amount of oil (or water) displaced by the duck, which we could use to calculate buoyancy.

2. The density of the crude oil. Different types of crude oil can have different densities, and the density would affect the buoyancy.

3. The density of the duck. In order to figure out the weight of the duck, we would need to know both its volume and its density. This would allow us to figure out when the duck's weight equals the buoyancy force in the oil.\\
\hline
\sethlcolor{orange}
\hl{Prompt:} Can you make some reasonable estimates and solve the problem?\\
\hline
\sethlcolor{gpt4}
\hl{ChatGPT:}
To make some reasonable estimates, let's say the following:

1. The volume of the duck: This can vary, but we can estimate an average duck to have a volume of 1 liter or 0.001 cubic meter.

2. The density of the crude oil: The density of crude oil varies based on its specific type and temperature, but a common value is around \( 850 \, \text{kg/m}^3 \).

3. The density of the duck: Most birds have a density slightly less than water due to their feathers and air-filled bones. Let's estimate it at about \( 950 \, \text{kg/m}^3 \)
, slightly less than the density of sea water which is about \( 1025 \, \text{kg/m}^3 \).\\
\hline
\end{tabular}}
\end{table}
\par ChatGPT was successful at filling in well-known factual information that was not provided in the problem statement, such as the speed of light in a vacuum, the speed of sound in water, the density of seawater, and even the height of the Salesforce Tower in San Francisco. However, its success rate dropped when the missing data involves more nuanced real-world understanding. In the Floating Duck problem presented above (Table~\ref{tab:follow-up}), ChatGPT made an incorrect assumption about the density of a duck being \( 950 \, \text{kg/m}^3 \) even after receiving the prompt to make reasonable estimates. A more accurate density value can be inferred from the information that 20\% of the duck's body is submerged in salt water, making its average density approximately \( 260\, \text{kg/m}^3 \).

\par Table~\ref{tab:coffee_mug} presents another problem that illustrates this failure mode. The problem asks a problem solver to estimate how long a driver should take to come to a stop while driving at 10 mph, so that a coffee cup that was accidentally left on the car's roof won't fly off. In analyzing ChatGPT's solution to this problem, we found that the model failed to provide a reasonable value for the friction coefficient between the coffee cup and the roof of the car (\(\mu\)). While the typical ceramic-on-metal and metal-on-metal static coefficients of friction are between 0.4 and 0.6, ChatGPT assumed the friction coefficient to be 1, as implied by its equating acceleration to the gravitational constant \textit{g} in the solution (highlighted in Table~\ref{tab:coffee_mug}). This incorrect assumption led to an erroneous value of the horizontal acceleration, which in turn resulted in an underestimate of the time it takes to stop the car safely. 

\begin{table}[htp]
\centering
\caption{The Coffee Cup on Car problem and ChatGPT's solution }
\label{tab:coffee_mug}
{\renewcommand{\arraystretch}{1.5}
\begin{tabular}{p{6cm}|p{9cm}}
\hline
\textbf{The Coffee Cup on Car Problem} & \textbf{The First Part of ChatGPT's Solution} \\
\hline
\sethlcolor{orange}
\hl{Prompt:}
Solve the following physics problem: You are driving your electric car to work one morning and forget your coffee cup on the roof. As you leave your driveway, you speed up to 10 mph and then remember your coffee cup and decide you need to stop. If you stop too fast, the cup could go flying and hit a pedestrian. How long should you take to come to a stop to be sure the cup doesn’t fly off the roof? & 

\sethlcolor{gpt4} \hl{ChatGPT:} 
\sethlcolor{yellow} To solve this problem, we need to consider the forces acting on the coffee cup, primarily gravity and friction. The friction force is given by the equation: \( F_{\text{friction}} = \mu N \), where \(\mu\) is the coefficient of friction and \(N\) is the normal force.

However, we don't have a specific value for the coefficient of friction between the cup and the roof of the car. \hl{Let's assume that the coffee cup will stay on the roof if the car stops with an acceleration equal to or less than g (the acceleration due to gravity)}, which is approximately 9.81 m/s².
\\
\hline
\end{tabular}}
\end{table}

\par To its credit, ChatGPT did hedge on its solution by making the following statement at the end: 
\begin{quote}
\sethlcolor{gpt4} \hl{ChatGPT:} Please note that this calculation assumes that the friction between the cup and the roof is strong enough to keep the cup from sliding when the car is decelerating at a rate equal to g. The actual time could be much longer if the friction is weaker than assumed. This is a simplified model and doesn't account for all factors that might be present in a real-world situation.
\end{quote}
This acknowledgement indicates that ChatGPT recognizes the potential error in it solution introduced by inaccurate assumptions. Nonetheless, we should be aware of its limitations, that it may not make accurate assumptions about missing information from the outset. Such difficulty in making reasonable estimates for real-world quantities is also typical of novice students.  

\par Another example that falls into this failure mode is ChatGPT's solution to a problem that involves estimating how the velocity of a merry-go-around would change once a person hops on after first pushing it around for three seconds. As the problem does not specify the initial speed of the merry-go-around, ChatGPT first estimated it to be around 14 mph through calculation. However, this estimate substantially exceeds the safe speed for merry-go-arounds, which are typically around 9 mph as stipulated by the US Consumer Product Commission. This example further illustrates that ChatGPT cannot always align and verify computational results with real-world standards and constraints. ChatGPT's solutions to eight problems had this failure mode.

\subsubsection{Calculation error}
While the first two failure modes are mostly associated with under-specified problems, the last failure mode, calculation error, has been observed in both well-specified and under-specified problems in the dataset. A total of six incorrect solutions, four for well-specified problems and two for under-specified problems, had this failure mode. The errors range from arithmetic to trigonometry, and underscore ChatGPT's well-known difficulties with mathematical computations. 

\subsubsection{An idiosyncratic solution case}
\par In addition to the three failure modes discussed above, we also identified an idiosyncratic case where ChatGPT reached the correct answer (therefore coded as correctly-solved) while disregarding the data provided in the problem. The problem is a well-specified one and involves the conversion of horsepower to kilowatts (Table~\ref{tab:horsepower}). Despite being provided with data in the problem statement, ChatGPT opted to utilize different data, that one horsepower is defined as the ability to lift 550 pounds one foot in one second, for its calculations. This behavior raises questions regarding how the underlying GPT-4 model potentially prioritizes its training data over new information in problem-solving. 

\begin{table}[h]
\centering
\caption{The problem for which ChatGPT did not use the data provided}
\label{tab:horsepower}
{\renewcommand{\arraystretch}{1.5}
\begin{tabular}{p{16cm}}
\hline
\multicolumn{1}{c}{\textbf{The Horsepower Problem}}\\
\hline
Engine power is sometimes expressed in terms of “horsepower.” One horsepower was defined by James Watt, who observed that a horse could turn a mill wheel with a radius of 12 ft at a constant rate of 144 times per hour, exerting a nearly constant force of 800 N tangentially to the wheel. Derive the conversion for horsepower to kilowatts \\
\hline
\end{tabular}}
\end{table}

\subsection{Chain-of-thought Prompt Engineering}
To what extent did prompt engineering enhance ChatGPT's problem-solving performance? In the second experiment, we applied the "solve the following physics problem step-by-step" prompt to all 40 problems in the dataset. Among the 12 problems that ChatGPT initially solved correctly, it generated consistent correct solutions for 11 of them under prompt engineering. However, ChatGPT made a calculation error involving trigonometry in one of the problems. Interestingly, in the idiosyncratic case where ChatGPT reached the correct answer without utilizing the given data in its initial solution, the step-by-step prompt helped it incorporate the data provided in the problem statement in its calculation for the correct solution.

\par Among the 28 problems that ChatGPT initially failed to solve, it was able to correctly solve three with the step-by-step prompting. Two of the three were related to the failure mode of ChatGPT not being able to construct accurate models about the real world. For the first one, ChatGPT initially did not subtract the weight of the water from a squid when it ejected water to create a form of jet-propulsion. For the other one, it initially treated a marble as a non-rotating block and did not account for the rotational kinetic energy as it rolled up a ramp. The last one of the three was associated with the initial failure mode of not being able to make reasonable assumptions about missing data, in this case the coefficient of friction in the Coffee Cup on Car problem. Table~\ref{tab:coffee_mug_solution_pe} presents ChatGPT's updated solution. The prompt of solving the problem step-by-step led to more precise and deliberate problem-solving as illustrated in this example. ChatGPT first broke down the solution process into discrete steps, then noted that without knowing the coefficient of static friction, a specific numerical answer could not be provided. After receiving a follow-up instruction through the dialogue input box to make reasonable assumptions, it chose a reasonable value of 0.6 for the friction coefficient and successfully solved the problem, unlike what it did in the absence of prompt engineering.

\par Overall, the results suggest that prompt engineering had a moderate effect on enhancing ChatGPT's problem-solving performance by constructing accurate models of the problem and making reasonable assumptions, though this effect is not statistically significant (Chi-squared (1) = 0.06, \textit{p} = 0.81). It should also be noted that step-by-step prompts had no impact on reducing calculation errors. 

\begin{table}[htp]
    \centering
    \caption{ChatGPT's updated solution to the Coffee Cup on Car problem after prompt engineering}
    \label{tab:coffee_mug_solution_pe}
    {\renewcommand{\arraystretch}{1.5}
    \begin{tabular}{p{0.95\linewidth}}
        \hline
        \sethlcolor{orange}
        \hl{Prompt:} Solve the following physics problem step-by-step. You are driving your electric car to work one morning and forget your coffee cup on the roof. As you leave your driveway, you speed up to 10 mph and then remember your coffee cup and decide you need to stop. If you stop too fast, the cup could go flying and hit a pedestrian. How long should you take to come to a stop to be sure the cup doesn’t fly off the roof?\\

        \hline
        \sethlcolor{gpt4}
        \hl{ChatGPT:}
        We can analyze this scenario by considering the forces acting on the coffee cup. Since the car is accelerating in the opposite direction of its motion (decelerating), there will be a frictional force trying to keep the cup in place. Let's assume the coefficient of static friction between the cup and the car's roof is \( \mu_s \), and that this is the only force preventing the cup from sliding.\\
        
        \textbf{1. Find the Maximum Deceleration:}\\
           The force of static friction is given by:
           \[
           F_{\text{friction}} = \mu_s \cdot F_n
           \]
           where \( F_n \) is the normal force, which is equal to the weight of the coffee cup:
           \[
           F_n = m \cdot g
           \]
           Here, \(m\) is the mass of the cup, and \(g\) is the acceleration due to gravity (\(9.8\, \text{m/s}^2\)
        ). The maximum static frictional force before the cup starts sliding is:
           \[
           F_{\text{max}} = \mu_s \cdot m \cdot g
           \]
        
        \textbf{2. Calculate the Deceleration:}
           This frictional force provides a maximum deceleration the car can undergo without the cup sliding. Using Newton's second law, \( F = ma \):
           \[
           \mu_s \cdot m \cdot g = m \cdot a
           \]
           Solving for \( a \), we find:
           \[
           a = \mu_s \cdot g
           \]
        
        \textbf{3. Find the Time to Stop:}
           Now we have the maximum deceleration, we can calculate the time it will take to stop the car from 10 mph under this deceleration. Using the equations of motion:
           \[
           v_f = v_i + at
           \]
           where \( v_f = 0 \, \text{m/s} \) (final velocity), \( v_i = 10 \, \text{mph} = 4.47 \, \text{m/s} \) (initial velocity), and \( a = -\mu_s \cdot g \) (deceleration). Solving for \( t \):
           \[
           t = \frac{v_f - v_i}{a} = \frac{-4.47}{-\mu_s \cdot 9.8}
           \]
           
        We need the coefficient of static friction to calculate the exact value of \( t \), which will tell us how long the driver should take to come to a stop to ensure the coffee cup doesn’t fly off the roof. If we had that value, we could plug it into our equation to find the time. In real-world scenarios, the coefficient of static friction might vary widely depending on the materials in contact and the conditions, so without specific information, it's hard to give a precise time.\\
        \hline
        \sethlcolor{orange}
        \hl{Prompt:}
        Please make reasonable assumptions about the coefficient of friction and solve the problem. \\
        \hline
        \sethlcolor{gpt4}
        \hl{ChatGPT:}    
        Certainly! To solve the problem, we'll make a reasonable estimate for the coefficient of static friction between the coffee cup and the car's roof.\\

        In general, the coefficient of static friction for smooth surfaces like a car roof and a typical coffee cup might range from 0.5 to 0.8. However, since a car roof is usually not designed to hold objects and the bottom of a coffee cup may be somewhat smooth, we'll use a more conservative estimate of \( \mu_s = 0.6 \).
        
        Now, we'll use our earlier formula for the time taken to stop:
        
        \[ t = \frac{-4.47}{-\mu_s \cdot 9.8} \]
        
        Substituting \( \mu_s = 0.6 \):
        
        \[ t = \frac{-4.47}{-0.6 \times 9.8} \approx 0.76\, \text{seconds} \]
        
        So the driver should take approximately 0.76 seconds to come to a stop to ensure that the coffee cup doesn't fly off the roof. Of course, this is a simplified model, and in a real-world scenario, other factors such as wind resistance, the curvature of the roof, and the exact materials of the cup and roof would also influence the outcome.\\
    \hline
    \end{tabular}}
\end{table}

\section{Discussion}
The present study found a marked difference in ChatGPT's problem-solving performance between well-specified and under-specified problems. The problems used in the study are all situated in real-world contexts and require the application of physics knowledge, yet differ in how much information is specified in the problem statement. ChatGPT performed better in well-specified problem, although it made occasional calculation errors. In contrast, it was far less accurate in solving under-specified problems. Two specific failure modes were observed: the first one being failure to construct accurate models of the physical world and reason about relationships between different variables in a model, and the second one being failure to make reasonable estimates or assumptions about the missing data. Prompt engineering produced a moderate improvement in ChatGPT's problem-solving performance. The prompt of solving a problem step-by-step proved moderately beneficial in guiding the AI model to be more deliberate and accurate in estimating missing data and constructing models of the problems, though it did not alleviate calculation errors. 

\subsection{Implications for Education}
The problem-solving process adopted by experts in science and engineering domains can be characterized as a series of interlinked decisions \citep{price2021detailed}. Utilizing this framework to analyze ChatGPT's performance on solving problems situated in real-world contexts, we note that the ChatGPT (based on GPT-4 model) demonstrated proficiency in deciding on the relevant domain-specific concepts and formulas based on a problem statement. At the same time, it fell short in making several key decisions, including determining how to construct a suitable model of a problem, and deciding how to make reasonable assumptions or estimates about incomplete data. 

\par These results have significant implications for STEM education, especially around how to leverage LLM-based tools like ChatGPT to help students develop expertise in problem-solving. First, the study identified facets of problem-solving where ChatGPT is indeed effective, namely identifying the relevant physics concepts needed for solving a problem based on the problem statement. This opens the possibility for ChatGPT to serve as a tutor for domain-specific problems and support students to pinpoint the essential knowledge underlying each problem and enhance their understanding of conceptual knowledge through problem-solving. This tutoring capability is particularly important as students struggle to decide on relevant physics concepts and formulas through analyzing the problem's statement, instead they rely on ineffective strategies such as searching for equations that contain the same variables to solve problems\cite{burkholder2020template, PhysRevSTPER.5.020102}. Given the capability of ChatGPT in deciding on relevant concepts, students can query ChatGPT with prompts such as "identify the relevant concepts associated with the following problem."  However, one concern associated with this use case is that ChatGPT may generate articulate, plausible-sounding, yet incorrect solutions based on the identified concepts. This presents a risk of misleading students and inducing misconceptions. Therefore, it is crucial to educate students on the problem-solving capabilities of ChatGPT (e.g., identifying the relevant concepts) as well as its shortcomings (e.g., generating inaccurate solutions due to failure to construct accurate models, failure to make reasonable assumptions, or calculation errors,  particularly in the case of under-specified problems).

\par Second, the findings point to what we should prioritize in STEM education in an era of increasingly powerful AI technologies. To prepare students for solving authentic problems in their professional and personal lives, STEM courses must place an emphasis on fostering effective decision-making practices. Specifically, students must have opportunities to practice making decisions related to construct appropriate models based on complex, real-world scenarios, as well as practice making decisions on what data is needed for solving a given problem, how to collect the data, and how to critically evaluate data quality. Mastery in these decisions will help students decompose complex, under-specified real-world challenges into a series of tractable, well-specified sub-problems for AI tools like ChatGPT to solve. The emphasis on developing problem-solving and decision-making expertise aligns well with the broader educational goal of preparing students to navigate a future of human-AI collaboration. 

Lastly, our findings have immediate implications for how to design homework and exam problems that are resilient to automatic solving by tools like ChatGPT. The key strategy involves incorporating authentic problems into teaching and assessment materials. These problems are not solvable by ChatGPT alone, and necessitate students to make informed decisions on how to utilize ChatGPT as a tool. At the same time, students must remain actively involved in constructing accurate models in real-world contexts and handle under-specified information. The inclusion of such authentic problems allows for a more valid assessment of student competencies in STEM courses.

\subsection{Implications for Human-AI Collaboration}
This study also provides insights for the future of human-AI collaboration. While LLMs like GPT-4 can solve well-specified problems, albeit with occasional calculation errors, human intervention and judgment is needed for navigating the complexity and ambiguity associated with authentic problems. This insight suggests a complementary relationship between human intelligence and artificial intelligence in addressing complex, authentic problems in the real world. Specifically, human experience and expertise can help construct accurate models of the physical world and make reasonable estimates or data collection plans for missing information. At the same time, AI's computational capability to instantly sift through vast knowledge bases and pinpoint the relevant domain knowledge constitutes an important asset to support human problem-solving.  

\section{Limitations}
In evaluating ChatGPT's capacity for problem-solving, it is important to recognize the inherent limitations associated with the underlying algorithm's probabilistic nature. ChatGPT may generate different answers each time a problem is posed, and this variability presents a challenge in our analysis of its solutions. The different releases and incremental builds of the algorithm could further produce varied results. Therefore, the interpretation of our findings must consider the specific version of the algorithm utilized, which spans from May to August 2023. Additionally, the current study did not ask ChatGPT to generate solutions for identical problems and prompts multiple times. This absence of repetitive testing restricts our understanding of the tool's stability and reliability in providing consistent solutions. The probabilistic and evolving nature of LLMs underscore the need for continuous evaluation and validation of their problem-solving capabilities in future studies.  

\section{Conclusion}
This study probed the capabilities and limitations of LLM-based technologies such as ChatGPT in solving authentic problems that are situated in real-world contexts and under-specified in terms of the requisite data. By focusing on the domain of physics, we were able to incorporate a diverse set of real-world scenarios into the problem set. The problem-solving practices and processes adopted to solve these physics problems are also applicable in the broader fields of science and engineering. Furthermore, the decision to include problems from well-specified to under-specified in terms of the amount of information provided in the problem statement led to a nuanced understanding of ChatGPT's capacity for solving different types of problems. The findings revealed that ChatGPT is adept at identifying relevant physics knowledge and applying it to solve well-specified problems. At the same time, its performance is less robust in modeling real-world complexities and making reasonable assumptions when data is missing in under-specified problems. 

\par These findings lead to future studies to investigate how LLMs can be utilized in STEM education to support student learning, such as serving as personalized tutors to scaffold students in identifying the relevant knowledge for solving a problem. Additionally, the insights from this study shed light on what are the key competencies for students to develop to prepare for a future where AI can effectively address well-specified problems. These competencies include the ability to construct accurate and concise models of problems, make deliberate decisions regarding assumptions and estimates, and devise plans for data collection. Students' mastery of these competencies, in conjunction with the advancement of AI technologies, pave the way for a future where human-AI collaboration can effectively address complex challenges in the real world.

\clearpage
\bibliographystyle{unsrt}  
\bibliography{references}  
\clearpage

\begin{landscape}
\appendix
\section{Appendix I - All problems used in the study grouped by data specificity}
\label{app:problem_details}
    \begin{longtable}{p{0.22\textwidth} p{0.25\textwidth}  p{0.1\textwidth} p{0.35\textwidth} p{0.15\textwidth}}
        \toprule
       \textbf{Problem} & \textbf{Concepts}  &\textbf{Solution} & \textbf{Failure Mode} & \textbf{Prompt \newline Engineering }\\
        \toprule
        \endfirsthead
        
        \toprule
        \textbf{Problem} & \textbf{Concepts}  &\textbf{ Solution} & \textbf{Failure Mode} & \textbf{Prompt \newline Engineering }\\
        \toprule        
        \endhead

        \multicolumn{5}{l}{Well-specified Problems} \\
        \midrule
        Hanging Sculpture & Force  & Incorrect & Calculation error & Incorrect \\
        Car Collision & Linear motion  & Incorrect & Calculation error & Incorrect \\

        Truck Hitch & Linear motion & Incorrect & Calculation error & Incorrect \\
        Sump Pump & Fluid mechanics & Incorrect & Calculation error & Incorrect \\
        Spinning Ball & Angular motion  & Incorrect & Failure to construct accurate models about the physical world & Incorrect \\
        Marble Rolling & Kinetic and potential energy & Incorrect & Failure to construct accurate models about the physical world & Correct \\
        Horsepower & Power  & Correct & Arrived at the correct solution but did not use the data provided & Correct \\
        Pendulum & Kinetic and potential energy  & Correct & & Incorrect \\
        Log Cabin & Force  & Correct & & Correct \\
        Breaking Bone & Linear motion  & Correct & & Correct \\
        Skiing off a Slope & Linear motion  & Correct & & Correct \\
        Athlete Cycling & Power  & Correct & & Correct \\
        Reciprocating Motion & Angular and linear motion  & Correct & & Correct \\
        Piano Wire Tension & Harmonic motion & Correct & & Correct \\
        Solar Panel & Energy & Correct & & Correct \\
        U-tube Manometer & Fluid mechanics & Correct & & Correct \\

        \midrule
        \multicolumn{5}{l}{Under-specified Problems} \\
        \midrule
        Millennium Tower & Force  & Incorrect & Failure to construct accurate models about the physical world & Incorrect \\
        Circular Table & Torque & Incorrect & Failure to construct accurate models about the physical world & Incorrect \\
        Dresser Tip-over & Torque  & Incorrect & Failure to construct accurate models about the physical world & Incorrect \\
        Pulling Boat & Torque  & Incorrect & Failure to construct accurate models about the physical world & Incorrect \\
        Bart Seat & Torque  & Incorrect & Failure to construct accurate models about the physical world & Incorrect \\
  
        Squid Movement II & Linear motion  & Incorrect & Failure to construct accurate models about the physical world & Incorrect \\
        Salesforce Tower Elevator & Linear motion  & Incorrect & Failure to construct accurate models about the physical world & Incorrect \\
        Dog Jumping & Linear motion  & Incorrect & Failure to construct accurate models about the physical world & Incorrect \\
        Rotor Ride & Force and angular motion  & Incorrect & Failure to make reasonable assumptions about missing data & Incorrect \\
        Ferris Wheel & Force and angular motion  & Incorrect & Failure to make reasonable assumptions about missing data & Incorrect \\
        Fishing & Angular motion  & Incorrect & Failure to make reasonable assumptions about missing data & Incorrect \\
        Race Car & Angular motion  & Incorrect & Failure to make reasonable assumptions about missing data & Incorrect \\
        Merry-go-round & Angular motion  & Incorrect & Failure to make reasonable assumptions about missing data & Incorrect \\
        Car Brakes & Angular and linear motion  & Incorrect & Failure to construct accurate models about the physical world & Incorrect \\
        Climbing Hill & Kinetic and potential energy & Incorrect & Calculation error & Incorrect \\
        Truck downhill & Kinetic and potential energy & Incorrect & Failure to construct accurate models about the physical world & Incorrect \\
        
        Helmet Protection & Kinetic energy  & Incorrect & Failure to make reasonable assumptions about missing data & Incorrect \\
        
        Floating Duck & Fluid mechanics &  Incorrect & Failure to make reasonable assumptions about missing data & Incorrect \\
        
        Shock Wave & Fluid mechanics  & Incorrect & Calculation error & Incorrect \\
        Diving in Pool & Linear motion and fluid mechanics  & Incorrect & Failure to construct accurate models about the physical world & Incorrect \\
        Coffee Cup on Car & Linear motion  & Incorrect & Failure to make reasonable assumptions about missing data & Correct \\
        Squid Movement I & Linear motion  & Incorrect & Failure to construct accurate models about the physical world & Correct \\
        Pendulum of Clock & Harmonic motion & Correct & & Correct \\
        Radio Station Wavelength & Wave mechanics & Correct & & Correct \\
        \bottomrule
    \end{longtable}
\end{landscape}

\end{document}